\documentclass[letterpaper]{article} 

\usepackage{subcaption}
\usepackage{amsmath}
\usepackage{multirow}
\usepackage{amssymb}
\usepackage[table]{xcolor}

\usepackage{aaai24}  
\usepackage{times}  
\usepackage{helvet}  
\usepackage{courier}  
\usepackage[hyphens]{url}  
\usepackage{graphicx} 
\usepackage{natbib}  
\usepackage{caption} 
\usepackage{algorithm}
\usepackage{algorithmic}

\usepackage{newfloat}
\usepackage{listings}
\DeclareCaptionStyle{ruled}{labelfont=normalfont,labelsep=colon,strut=off} 
\floatstyle{ruled}
\newfloat{listing}{tb}{lst}{}
\floatname{listing}{Listing}
\title{BaCon: Boosting Imbalanced Semi-supervised Learning via Balanced Feature-Level Contrastive Learning}
\author{
    Qianhan Feng\textsuperscript{\rm 1,2},
    Lujing Xie\textsuperscript{\rm 3},
    Shijie Fang\textsuperscript{\rm 2,4}\thanks{Work done as a master student at Peking University.},
    Tong Lin\textsuperscript{\rm 1,2}\thanks{Corresponding Aurthor.}\\
}
\affiliations{
    \textsuperscript{\rm 1}National Key Laboratory of General Artificial Intelligence, China\\
    \textsuperscript{\rm 2}School of Intelligence Science and Technology, Peking University\\
    \textsuperscript{\rm 3}Yuanpei College, Peking University\\
    \textsuperscript{\rm 4}Google, Shanghai, China\\

    \{fengqianhan, lujing\_xie\}@stu.pku.edu.cn, shijiefang@google.com, lintong@pku.edu.cn
}

\usepackage{bibentry}

\begin{document}

\maketitle

\begin{abstract}
Semi-supervised Learning (SSL) reduces the need for extensive annotations in deep learning, but the more realistic challenge of imbalanced data distribution in SSL remains largely unexplored. In Class Imbalanced Semi-supervised Learning (CISSL), the bias introduced by unreliable pseudo-labels can be exacerbated by imbalanced data distributions. Most existing methods address this issue at instance-level through reweighting or resampling, but the performance is heavily limited by their reliance on biased backbone representation. Some other methods do perform feature-level adjustments like feature blending but might introduce unfavorable noise. In this paper, we discuss the bonus of a more balanced feature distribution for the CISSL problem, and further propose a \textbf{Ba}lanced Feature-Level \textbf{Con}trastive Learning method (\textbf{BaCon}). Our method directly regularizes the distribution of instances' representations in a well-designed contrastive manner. Specifically, class-wise feature centers are computed as the positive anchors, while negative anchors are selected by a straightforward yet effective mechanism. A distribution-related temperature adjustment is leveraged to control the class-wise contrastive degrees dynamically. Our method demonstrates its effectiveness through comprehensive experiments on the CIFAR10-LT, CIFAR100-LT, STL10-LT, and SVHN-LT datasets across various settings. For example, BaCon surpasses instance-level method FixMatch-based ABC on CIFAR10-LT with a 1.21\% accuracy improvement, and outperforms state-of-the-art feature-level method CoSSL on CIFAR100-LT with a 0.63\% accuracy improvement. When encountering more extreme imbalance degree, BaCon also shows better robustness than other methods.  
\end{abstract}

\section{Introduction}

    Recently, several Semi-supervised Learning (SSL) methods have been proposed to alleviate the burden of time-consuming data labeling. Most of these methods incorporate unlabeled samples into model training using consistency constraints and pseudo-labeling \cite{DBLP:conf/nips/SohnBCZZRCKL20,DBLP:conf/nips/ZhangWHWWOS21}. However, these methods are often studied under the assumption of equally distributed unlabeled data, which may not hold true in realistic scenarios. Although some Class Imbalanced Semi-supervised Learning (CISSL) methods have been proposed, most of them resample and redistribute at the instance level, ignoring the upstream representation that has a significant impact. Even if some methods \cite{fan2022cossl} make up for this limitation with feature blending from the perspective of representation, it may not only bring more feature noise but also be subject to the spurious prior knowledge of the same distribution of labeled data and unlabeled data. The issue of biased feature representation in the CISSL setting has yet to be thoroughly explored.   

\begin{figure}[t]
    \centering
    \includegraphics[width=1.0\columnwidth]{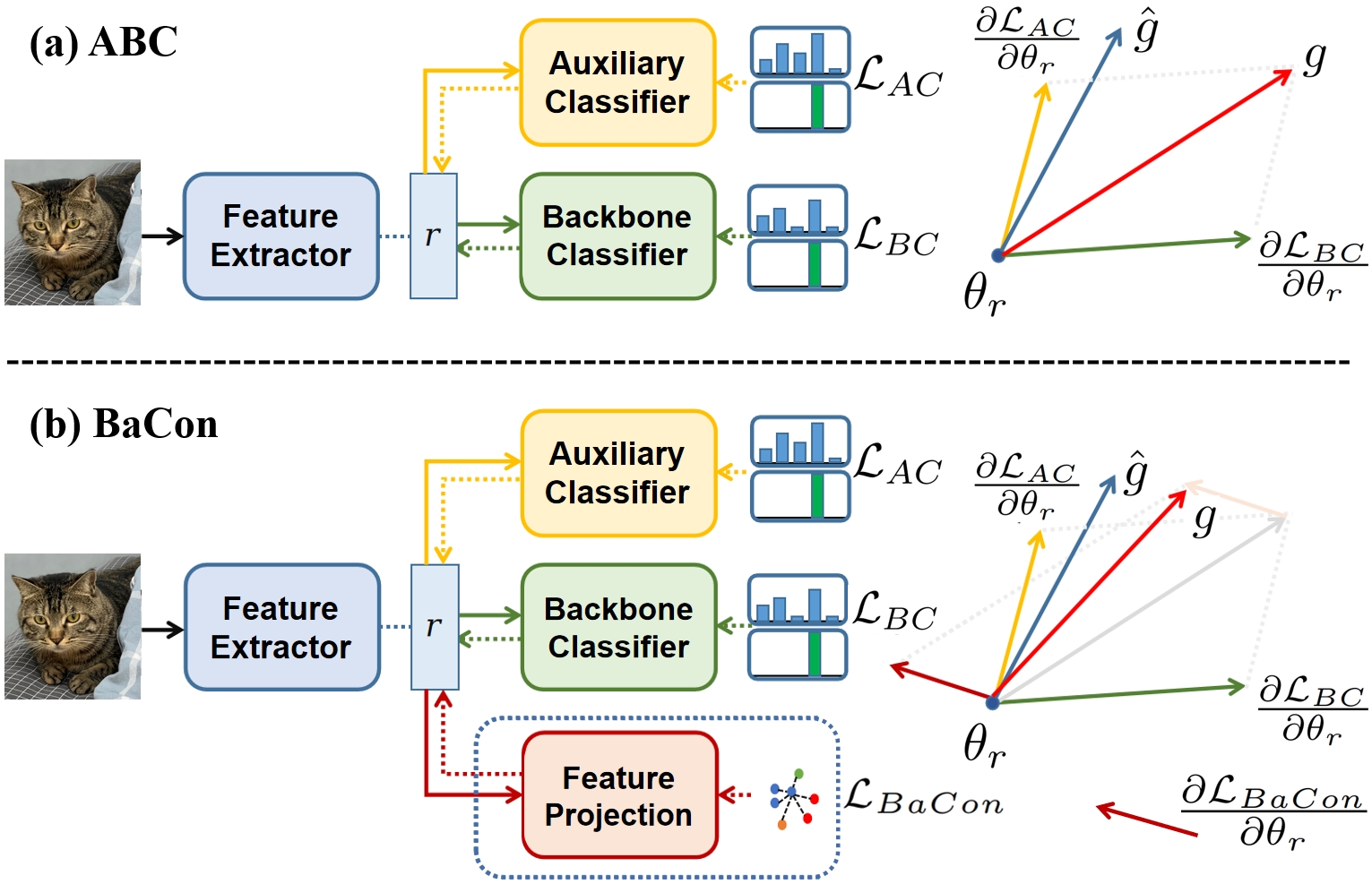}
    \caption{Gradient of ABC on representation layer is still biased. In contrast, BaCon provides an extra gradient that narrows the gap between the estimated gradient $g$ and the ideal optimal gradient $\hat{g}$. $r$ is the feature representation.}
    \label{fig:headfig}
\end{figure}

In this paper, we first examine the limitations of the state-of-the-art instance-level CISSL method called ABC (Auxiliary Balanced Classifier). ABC adds an auxiliary classification head onto the backbone classifier, and uses inversely proportional Bernoulli mask to achieve balanced learning. There are two sources of gradients that propagate back into the upstream feature extractor: one from the original backbone classifier and the other from the auxiliary classifier. As shown in Figure \ref{fig:headfig}(a), the effect of the combination of these two gradients on the representation layer is still very different from the ideal optimal gradient direction. This would result in the uneven distribution of representations, and directly leads to the degraded performance of the overall CISSL method. 

Then, we propose to solve this problem from the aspect of feature distribution, and add an additional contrastive loss to directly perform distribution regularization of the feature representations provided by the feature extractor (Figure \ref{fig:headfig}(b)). To be more specific, a projection head is firstly used to map the representation $r$ into another contrastive space, and then the corresponding features of reliable instances are recorded. To implement global positive contrastive learning, the feature centers of each category are calculated to act as the positive anchors in the proposed contrastive loss. Furthermore, a Reliable Negative Selection method (RNS) is designed to easily find adequate and reliable negative samples within a mini-batch for contrastive learning. 

However, simply applying equal degree of contrast to each category of imbalanced data sizes cannot fully achieve the desired balanced feature distribution. To tackle this issue, a Balanced Temperature Adjusting mechanism (BTA) is proposed to dynamically adjust the temperature coefficient in the contrastive loss according to the fluctuant distribution, achieving self-adaptive learning. To sum up, our work has the following contributions: 
\begin{itemize}
\item[$\bullet$]  We discuss the limitations of previous instance-level CISSL methods from the perspective of representation distribution. Based on this, we propose a contrastive learning method to directly regularize feature-level distribution.  
\item[$\bullet$]  We design an easy and reasonable way to construct positives and negatives for contrastive learning. And a self-adaptive mechanism is proposed to adjust the class-wise learning degree based on the imbalanced distribution.  
\item[$\bullet$]  Extensive experiments across various datasets and settings demonstrate the effectiveness of our method.
\end{itemize}

\section{Related Work}
\subsection{Semi-supervised Learning}

Semi-supervised Learning has a long history of research \cite{zhu2005semi, ouali2020overview}. In deep learning, many SSL algorithms have been proposed under the paradigms of mean teacher\cite{tarvainen2017mean}, pseudo labeling \cite{lee2013pseudo} and consistency regularization \cite{berthelot2019mixmatch,DBLP:conf/nips/XieDHL020,DBLP:conf/iclr/Wang0HHFW0SSRS023}. ReMixMatch \cite{DBLP:conf/iclr/BerthelotCCKSZR20} introduces Distribution Alignment, which enforces that the aggregate of predictions on unlabeled data matches the distribution of the provided labeled data.
In FixMatch \citep{DBLP:conf/nips/SohnBCZZRCKL20}, the weakly-augmented unlabeled example is first fed to the model to obtain the reliable pseudo-label. Then consistency regularization is performed between the prediction of the strongly-augmented version and the pseudo label. FlexMatch \citep{DBLP:conf/nips/ZhangWHWWOS21} further proposes an adaptive threshold mechanism according to the different learning stages and categories. SimMatch \citep{zheng2022simmatch} considers semantic similarity and instance similarity simultaneously to encourage consistent prediction. Besides, explicit consistency regularization is also widely explored \citep{DBLP:conf/iclr/LaineA17,DBLP:journals/pami/MiyatoMKI19,ganev2020semi}.

However, these methods are designed under the balanced data distribution, and would fail miserably when encountering with imbalanced training data.

\subsection{Class Imbalanced Semi-supervised Learning}

Recent works have made great progress in addressing the issue of CISSL. DARP \cite{DBLP:conf/nips/KimHPYHS20} softly refines the pseudo-labels generated from the biased model by solving a convex optimization problem. Wei et al. \cite{DBLP:conf/cvpr/WeiSMYY21} observe that models typically exhibit high precision but low recall on minority classes, consequently proposing a reverse sampling methodology. The state-of-the-art instance-level method ABC \cite{DBLP:conf/nips/LeeSK21} utilizes an auxiliary classifier to rebalance the class distribution. Meanwhile, DASO \cite{DBLP:conf/cvpr/Oh0K22} establishes a similarity-based classifier and a linear classifier, blending the pseudo-label in accordance with the pseudo-label distribution. CoSSL \cite{fan2022cossl} decouples representation and classifier learning, increases the data diversity of minority classes by feature blending, but it brings a lot of noise and relies on the prior of similar distribution between labeled and unlabeled data. 

While most of these methods concentrate on instance-level design, it has been observed that incorporating a self-supervision framework can consistently enhance the final performance \cite{yang2020rethinking}. We argue that due to the fact that the classifier is biased towards the majority classes, the feature extractor may not learn high-quality balanced representations. Thus, we are inspired to further optimize the representations during training.

\subsection{Contrastive Learning}

Contrastive Learning, successful in self-supervised vision tasks using various strategies \cite{DBLP:journals/corr/abs-1807-03748,chen2020simple,tian2020contrastive,he2020momentum,caron2020unsupervised,grill2020bootstrap,chen2021exploring}. Particularly, the instance discrimination task \cite{wu2018unsupervised} with NCE loss \cite{gutmann2010noise} and InfoNCE Loss \cite{DBLP:journals/corr/abs-1807-03748} have led to milestone works \cite{he2020momentum, chen2020simple}. These works also propose techniques like momentum encoder, memory bank and projection head.


Despite leveraging novel and successful ideas in Self-supervised Learning, Semi-supervised Learning methods have limited access to labeled data. CoMatch \cite{li2021comatch} regularizes the structure of embeddings to smooth the class probabilities in a contrastive manner. U$^2$PL \cite{DBLP:conf/cvpr/WangWSFLJWZL22} makes use of the most reliable predictions to generate positive samples, while using the least reliable predictions to create a negative sample memory bank to address ambiguous boundaries in Semi-supervised Semantic Segmentation. With the help of labeled data, contrastive methods in semi-supervised field can be more natural and general, in contrast to considering a single example as a class in self-supervised methods.

\section{Proposed Method}
\subsection{Problem Setup}

For a K-class classification task, suppose we have a labeled dataset $\mathcal{X} =\left\{ \left ( x_{n},y_{n} \right ):n\in \left ( 1,...,N \right )\right\}$, where $x_{n}\in \mathbb{R}^{d}$ is the $n$-th labeled data sample, and $y_{n} \in \left\{1,...,K \right\}$ is the corresponding label. Additionally, an unlabeled dataset  $ \mathcal{U}= \left\{(u_{m}):m \in (1,...,M) \right\}$ is also provided, where $u_{m}\in \mathbb{R}^{d}$ is the $m$-th unlabeled instance. In SSL setting, only a few data points have accessible labels, while the rest are unlabeled. We denote the ratio of the amount of labeled data as $\beta = \frac{N}{M+N}$, which is generally small since data labeling is expensive. Then, the number of labeled data in the $k$-th class is denoted as $N_{k}$, and $\sum_{k}^{K}N_{k}=N$. Apart from the above setups, the imbalance degree across classes is also a key consideration. We assume that the $K$ classes are sorted in descending order, i.e. $N_{1}> N_{2}> \cdots >N_{K}$, and the imbalance degree of the dataset is represented by the imbalance ratio $\gamma=\frac{N_{1}}{N_{K}}$. We do not follow prior works in assuming that $\mathcal{X}$ and $\mathcal{U}$ share a similar distribution, this means that any extreme distributions can be possible. For example, the imbalance ratio of labeled and unlabeled data could be even inverse, i.e. $\gamma_{L}=1/\gamma_{U}$. Regarding the test time, we evaluate the effectiveness of our method on a class-balanced dataset.

\begin{figure}[t]
    \centering
    \includegraphics[width=0.9\columnwidth]{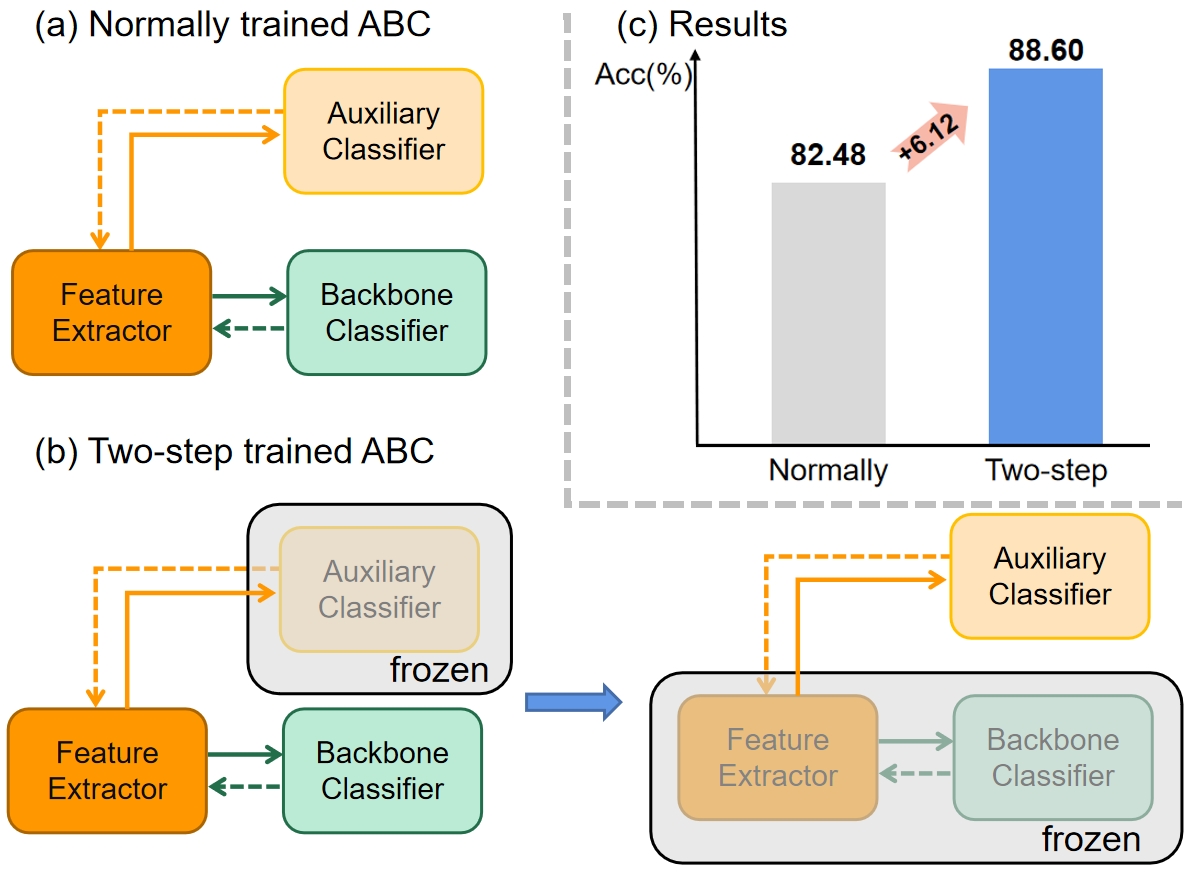}
    \caption{FixMatch-based ABC is trained in two steps as shown in (b)  as the comparison of the normally trained one in (a). Remarkable improvement can be observed in (c).}
    \label{fig:discussion}
\end{figure}

\subsection{Motivation}

Before introducing our method, we begin by discussing the limitations of the state-of-the-art instance-level approach ABC. ABC introduces an auxiliary classifier besides the one in backbone model, and generates an inversely proportional Bernoulli mask according to the predicted distribution. This mask is applied to the auxiliary classifier to provide a more balanced learning. However, the upstream feature extractor receives an imbalanced gradient from the backbone classifier head simultaneously, which may cause an annoying problem. This conflicting gradient pushes the representation learning away from the optimal gradient direction, thereby constraining the overall performance of ABC.

To investigate the potential benefits of a more balanced representation, we conduct a set of comparison experiments on the standard CIFAR10-LT dataset, which will be introduced in the experiment section. As a baseline, we simultaneously train the backbone model and the auxiliary classifier for 300,000 iterations. In the comparison setting, the backbone model is first trained on a balanced dataset with the same total number of samples as CIFAR10-LT for 300,000 iterations, aiming to learn a more balanced representation parameters. The auxiliary head is then trained on imbalanced CIFAR10-LT for another 300,000 iterations, using the frozen backbone parameters. 

The results of this exploration experiment are presented in Figure 2. It is obvious that when the auxiliary classifier learns on an imbalanced representation, only an accuracy of 82.48\% is achieved. As a comparison, the accuracy surges to 88.60\% when it learns on the frozen balanced representation layer. This intriguing outcome indicates that methods like ABC are considerably limited by the imbalanced learned representation. However, it is impossible to produce balanced instance distribution in practice as we did in former discussion. To overcome this challenge, we propose to directly regularize class-wise feature of the samples to have a more balanced distribution, thus facilitating downstream classification.

\subsection{Base SSL Algorithm} 

Our feature-level contrastive learning method can be viewed as a plug-in component, and we design to combine it with the state-of-the-art instance-level method ABC for better performance. ABC attaches an auxiliary classifier on backbone SSL algorithms like FixMatch and ReMixMatch, and we take FixMatch as an example here. Apart from the basic classification loss $\mathcal{L}_{S}$ calculated from the prediction of labeled data $\alpha(x_{b})$, FixMatch also uses the reliable prediction beyond threshold of weakly augmented version of unlabeled input $\alpha(u_{b})$ to produce pseudo label $\hat{q}_{b}$, in order to enable consistency regularization unsupervised loss $\mathcal{L}_{U}$ on the strongly augmented version of the same input $\mathcal{A}(u_{b})$. The $\mathcal{L}_{S}$ and $\mathcal{L}_{U}$ can be formulated as:
\begin{equation}
    \label{fix_suploss}
    \mathcal{L}_{S} = \frac{1}{B_{l}}\sum^{B_{l}}_{b=1}\mathbf{H}(p_{s}(y | \alpha(x_{b})),y_{b}) ,
\end{equation} 

\begin{equation}
    \label{fix_unsuploss}
    \mathcal{L}_{U} = \frac{1}{B_{u}}\sum^{B_{u}}_{b=1}\mathbf{H}(p_{s}(y | \mathcal{A}(x_{b})),\hat{q}_{b}),
\end{equation} 
where $B_{l}$ and $B_{u}$ are the number of labeled and unlabeled data within a mini-batch, and $\mathbf{H}$ is the cross-entropy loss, $p_{s}(y|\alpha(x_{b}))$ represents predicted confidence distribution of augmented labeled data $\alpha(x_{b})$.

The auxiliary classifier in ABC performs similar supervised and consistency-based unsupervised learning, the only difference is that a Bernoulli mask $\mathcal{M}$ inversely proportional to the predicted category size is applied. The classification loss of auxiliary head on labeled data $x_{b}$ is:
\begin{equation}
    \label{abcsup}
    \mathcal{L}_{cls}= \frac{1}{B_{l}}\sum^{B_{l}}_{b=1}\mathcal{M}(x_{b})\mathbf{H}(p_{a}(y|\alpha(x_{b})),y_{b}),
\end{equation} 

\begin{equation}
    \label{supbernoullimask}
    \mathcal{M}(x_{b})=\mathcal{B}\left(\frac{N_{L}}{N_{y_{b}}}\right).
\end{equation}  

The auxiliary head also makes prediction for weakly augmented version of unlabeled data $\alpha(u_{b})$ and strongly augmented version $\mathcal{A}(u_{b})$. $\mathcal{B}$ is the Bernoulli distribution generator.
\begin{equation}
\begin{split}
    \label{abcunsup}
    \mathcal{L}_{consis} & = 
    \frac{1}{B_{u}}\sum^{B_{u}}_{b=1} \\
     & \mathcal{M}(u_{b})\mathbf{I}(\max(q_{b})>\tau)\mathbf{H}(p_{a}(y | \mathcal{A}(x_{b})),\hat{q_{b}}) , 
\end{split}
\end{equation}
\begin{equation}
    \label{unsupbernoullimask}
    \mathcal{M}(u_{b})=\mathcal{B}\left(\frac{N_{L}}{N_{\hat{q}_{b}}}\right),
\end{equation}
where $\mathbf{I}$ is the indicator function, $max(q_{b})$ is the highest predicted assignment probability for any class, and $\tau$ is the confidence threshold set to 0.95 by default.

Finally, the total loss for backbone SSL algorithm with the auxiliary classifier is formulated as:
\begin{equation}
    \label{lossback}
    \mathcal{L}_{back}=\mathcal{L}_{S}+\mathcal{L}_{U}+\mathcal{L}_{cls}+\mathcal{L}_{consis}.
\end{equation} 

\begin{figure*}[htbp]
    \centering
    \includegraphics[width=1.9\columnwidth]{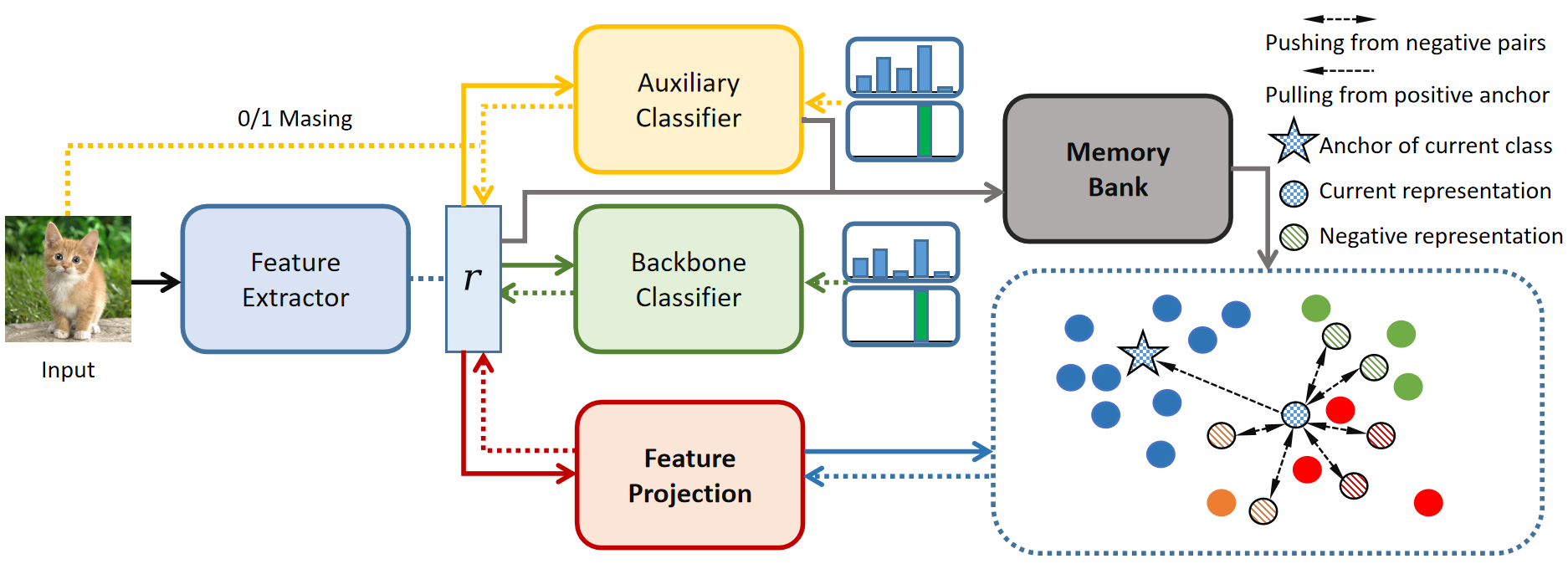}
    \caption{Overall training procedure of BaCon. The circle and star filled with squares represent the current representation and the corresponding positive anchor point, respectively. Circles filled with slashes in different colors represent negative instance features belonging to different classes in the current mini-batch.}
    \label{fig:main}
\end{figure*}

\subsection{Contrastive Learning} 

In the Motivation section, we have shown that a more balanced feature distribution can facilitate downstream classification task. Therefore, we propose to directly regularize the learned representation distribution using contrastive learning method.

The feature-level distribution regularization is achieved by a novel contrastive loss. Similar to InfoNCE Loss \cite{DBLP:journals/corr/abs-1807-03748}, we introduce the balanced feature-level contrastive loss, which is expressed as follows:

\begin{equation}
\begin{split}
\label{bacon}
\mathcal{L}_{BaCon} &= -\frac{1}{B}\sum^{K}_{k=1}\sum^{B_{k}}_{b=1}\\
    &\log\frac{e^{\left<f_{b},Anc_{k}\right>/\hat{\tau}}}{e^{\left<f_{b},Anc_{k}\right>/\hat{\tau}}+ \lambda\sum^{B_{\bar{k}}}_{q=1}e^{\left<f_{b},f_{q}\right>/\tau}},
\end{split}
\end{equation} 

where $B$ is the size of a mini-batch during training, $B_{k}$ is the number of samples predicted to belong to class $k$. Besides, $f_{b}$ and $Anc_{k}$ stand for the representation of current instance and the its positive anchor target. $\left<\cdot,\cdot\right>$ represents the cosine similarity function. $\hat{\tau}$ and $\tau$ are temperatures of positive and negative pairs. $\bar{k}$ stands for the subset in mini-batch that confidently does not belong to class $k$, and $\mathcal{B}_{\bar{k}}$ is its size while $f_{m}$ is the negative representation. $\lambda$ is the weight that controls the learning of negative samples. After presenting the loss, we begin to introduce details.

To get the representation of each input instance, we attach a linear projection head directly to the representation layer of the backbone. Feature of each instance is projected into another high-dimension space $\mathbb{D}$ where we perform the contrastive learning:  
\begin{equation}
\label{projection}
f_{b} = \mathcal{P}(\mathcal{F}(x_{b})), f \in \mathbb{D},
\end{equation} 
in which $\mathcal{P}$ is the linear projection.

After obtaining new representations, we maintain a memory bank $\mathcal{S}_{b}$ to store the features, which is updated every time when the data is sampled. In order to make the memory bank stable and reliable, only features of samples whose highest confidence score surpass the predefined threshold $\tau_{th}$ can be recorded:
\begin{equation}
\label{featureupdate}
f_{b}\mapsto\mathcal{S}_{b},\max(\sigma(\mathcal{H}_{A}(\mathcal{F}(x_{b})))>\tau_{th} ,
\end{equation}
where $\mathcal{H}_{A}$ is the auxiliary classifier and $\sigma$ is the $SoftMax$ function. $\tau_{th}$ is set to 0.98 by default.

Moreover, another memory bank $\mathcal{S}_{K}$ is also built to record samples' corresponding category, where the ground-truth label of labeled data and predicted result of unlabeled data from the auxiliary head is saved. Then, we calculate the feature center of each class by averaging representations according to the latest memory banks, taking the feature centers to be the anchor points $Anc_{k}$ which are the positive learning targets in $\mathcal{L}_{BaCon}$,
\begin{equation}
\label{featurecenter}
Anc_{k} = \frac{1}{N_{k}}\sum^{N_{k}}_{n=1}f_{n}, f_{n}\in \mathcal{S}_{b}\{k\} ,
\end{equation}
$\mathcal{S}_{b}\{k\}$ represents the subset in $\mathcal{S}_{b}$ that are predicted to belong to class $k$, and $N_{k}$ is the size of it.

\subsection{Reliable Negative Selection}

The selection of negative samples is challenging. Under the paradigm of pseudo labeling, if labeled instances are solely used for promoting the accuracy of negative samples, the size of negative batch would be reduced greatly. This could result in huge bias we intend to eliminate originally. On the other hand, if all samples that are not identified as current class are treated as negative samples for the sake of larger size of negatives, a lot of noise would be introduced, which will also make contrastive learning intractable. 

To tackle this dilemma, we propose the Reliable Negative Selection (RNS) scheme. In RNS, the representation of labeled data whose predicted confidence is above $\tau_{th}$ and belongs to other category rather than class $k$ is selected into reliable negatives.

As for the unlabeled data, RNS first sorts the output confidence scores by descending order and assigns an index $Idx(q)$ to the corresponding category $q$. Then, RNS searches for the representations of samples whose confidence score for class $k$ is outside of the top $n$ in the queue to be viewed as reliable unlabeled negatives, i.e. 
\begin{equation}
\label{sortunlabel}
f_{b}\mapsto \bar{k}, Idx(k)>n ,
\end{equation}
where $n$ is set to 3 by default.

In most SSL methods, many data points with low confidence scores are excluded from training, which in turn aggravates the lack of accessible data. Although samples with low confidence of class $k$ cannot directly provide information about what features of class $k$ should be like, they can indeed reliably provide key information about what the feature should not be like. With RNS, we reach a balance point where more data points are included and at the same time reliable contrastive information is also acquired.


However, the size of $\bar{k}$ can fluctuate significantly. To handle this problem, we make the weight $\lambda$ batchsize-related to stabilize the loss, thus the second term of denominator in Equation 
 (\ref{bacon}) can be formulated as:
\begin{equation}
\begin{split}
\label{lambda}
\lambda
    \sum^{B_{\bar{k}}}_{m=1}e^{\left<f_{b},f_{m}\right>/\tau}
=\frac{B}{B_{\bar{k}}}
\sum^{B_{\bar{k}}}_{m=1}e^{\left<f_{b},f_{m}\right>/\tau}.
\end{split}
\end{equation} 

\subsection{Balanced Temperature Adjusting} 

Although the attraction and the repelling in contrastive learning already provide a good regularization for feature distribution, we argue that the degree of class-wise clustering should be different. The attraction from the positive anchors of tail classes should be weaker for the following reason: the feature center might be biased from the ground-truth center because it is calculated by averaging on a small number of instances. Learning towards a less reliable target should be more careful.

With this reasonable analysis, we propose a Balanced distribution-related Temperature Adjusting method (BTA) to perform dynamic class-wise temperature modulation of positive pairs in $\mathcal{L}_{BaCon}$. The ratio of number of each category $N_{c}$ to the global maximum value $\max\{N_{C}\}$ is first calculated. In contrastive learning, temperature $\tau$ is used to scale the degree of learning. The smaller $\tau$ is, the greater the mutual attraction or repulsion would be. So we multiply it by a coefficient that is negatively correlated with the distribution ratio to dynamically control the contrastive learning.

In the later stage of training, the feature centers of different categories should be regionally stable. Therefore, the difference in temperature coefficients should be gradually reduced, so the final modulation can be formulated as:
\begin{equation}
\begin{split}
\label{finaltau}
\hat{\tau}_{c} = \tau\cdot\left[1-(1-\frac{t}{T})^2\cdot\sqrt{\frac{N_{c}}{\max\{N_{C}\}}}\cdot\eta\right], N_{c} \in \{N_{C}\} ,
\end{split}
\end{equation} 
where $\tau$ is the default temperature and $\eta$ controls the sensitivity of the temperature. $t$ is the current iteration number while $T$ is the total training iterations.

\subsection{Training and Inference} 

\begin{table*}[t!]
\centering
\begin{tabular}{c c c c c}
\hline \hline
  & CIFAR10-LT &  CIFAR100-LT & STL10-LT & SVHN-LT\\ 
\hline
 Method & $\gamma=100,\beta=20\%$ & $\gamma=20,\beta=40\%$ & $\gamma_{L}=10$ & $\gamma=100,\beta=20\%$ \\ 
\hline
Supervised Only & $57.16\pm0.35$ & $45.72\pm0.12$ & $45.23\pm0.23$ & $85.92\pm0.11$ \\
\hline
FixMatch\cite{DBLP:conf/nips/SohnBCZZRCKL20} & $75.30\pm0.37$ & $53.94\pm0.09$ & $67.16\pm0.36$  & $92.63\pm0.15$\\
w/ DASO\cite{DBLP:conf/cvpr/Oh0K22} & $74.78\pm0.21$ & $54.83\pm0.19$ & $68.69\pm0.15$ & $90.24\pm0.27$\\
w/ DebiasPL\cite{DBLP:conf/cvpr/WangWLY22} & $75.25\pm0.21$ & $54.90\pm0.11$ & $65.96\pm0.23$  & $92.42\pm0.24$\\
w/ CReST\cite{DBLP:conf/cvpr/WeiSMYY21} & $76.62\pm0.12$ & $54.83\pm0.10$ & $66.45\pm0.09$ & $93.25\pm0.07$\\
w/ CReST+PDA\cite{DBLP:conf/cvpr/WeiSMYY21} & $78.64\pm0.40$ & $55.01\pm0.12$ & $67.17\pm0.04$ & $93.23\pm0.12$\\
w/ DARP\cite{DBLP:conf/nips/KimHPYHS20} & $78.00\pm0.33$ & $55.63\pm0.07$ & $62.43\pm0.10$ & $92.49\pm0.16$\\
w/ Adsh\cite{DBLP:conf/icml/GuoL22} & $78.72\pm0.36$ & $53.97\pm0.12$ & $70.44\pm0.09$ & $92.71\pm0.13$\\
w/ SAW\cite{DBLP:conf/icml/LaiWGCC22} & $80.12\pm0.59$ & $55.87\pm0.05$ & $70.51\pm0.21$ & $92.92\pm0.09$\\
w/ ABC\cite{DBLP:conf/nips/LeeSK21} & $83.25\pm0.77$ & $56.91\pm0.02$ & $71.23\pm0.04$ & $94.15\pm0.04$\\
w/ CoSSL\cite{fan2022cossl} & $84.09\pm0.16$ & $57.33\pm0.05$ & $70.95\pm0.17$ & $93.39\pm0.05$\\
\rowcolor{lightgray} \textbf{w/ BaCon(Ours)} & $\textbf{84.46}\pm0.15$ & $\textbf{57.96}\pm0.26$ & $\textbf{71.55}\pm0.09$  & $\textbf{94.54}\pm0.06$ \\
\hline
ReMixMatch\cite{DBLP:conf/iclr/BerthelotCCKSZR20} & $77.96\pm0.24$ & $56.12\pm0.12$ & $66.97\pm0.04$  & $92.35\pm0.05$\\
w/ DASO\cite{DBLP:conf/cvpr/Oh0K22} & $78.86\pm0.15$ & $57.67\pm0.20$ & $65.38\pm0.18$  & $92.49\pm0.17$\\
w/ DebiasPL\cite{DBLP:conf/cvpr/WangWLY22} & $78.14\pm0.08$ & $56.85\pm0.19$ & $64.90\pm0.35$  & $92.41\pm0.09$\\
w/ CReST\cite{DBLP:conf/cvpr/WeiSMYY21} & $79.38\pm0.17$ & $59.14\pm0.11$ & $65.56\pm0.10$  & $93.63\pm0.06$\\
w/ CReST+PDA\cite{DBLP:conf/cvpr/WeiSMYY21} & $79.91\pm0.20$ & $59.78\pm0.23$ & $67.57\pm0.11$  & $93.74\pm0.15$\\
w/ DARP\cite{DBLP:conf/nips/KimHPYHS20} & $77.80\pm0.18$ & $57.21\pm0.21$ & $65.93\pm0.16$  & $92.47\pm0.04$\\
w/ SAW\cite{DBLP:conf/icml/LaiWGCC22} & $81.71\pm0.38$ & $32.53\pm0.67$ & $66.07\pm0.26$  & $93.42\pm0.51$\\
w/ ABC\cite{DBLP:conf/nips/LeeSK21} & $84.49\pm0.24$ & $59.92\pm0.01$ & $67.24\pm1.02$ & $94.03\pm0.18$ \\
w/ CoSSL\cite{fan2022cossl} & $84.93\pm0.02$ & $\textbf{60.46}\pm0.15$ & $68.73\pm0.77$  & $92.26\pm0.03$\\
\rowcolor{lightgray} \textbf{w/ BaCon(Ours)} & $\textbf{85.05}\pm0.09$ & $60.15\pm0.05$ & $\textbf{69.26}\pm0.83$ & $\textbf{94.35}\pm0.11$\\
\hline \hline
\end{tabular}
\caption{Overall results under different imbalance datasets with various semi-supervised learning algorithms. The results are reported according to balance accuracy(\%). Labeled data and unlabeled data share the same imbalance degree $\gamma$ in CIFAR10-LT, CIFAR100-LT and SVHN-LT datasets. But in STL10, only the imbalance ratio $\gamma_{L}$ of labeled data is available.}
\label{table:mainresults}
\end{table*}

During training, the backbone SSL algorithm and auxiliary classification head are trained first to warmup the memory banks, and the $\mathcal{L}_{BaCon}$ is added to the total loss after warmup. The total loss can be formulated as 
\begin{equation}
\begin{split}
\label{loss}
\mathcal{L}&=\mathcal{L}_{back}+\mathbf{1}(t)\cdot\mathcal{L}_{BaCon} \\
 &=\mathcal{L}_{S}+\mathcal{L}_{U}+\mathcal{L}_{cls}+\mathcal{L}_{consis}+\mathbf{1}(t)\cdot\mathcal{L}_{BaCon},
    \end{split}
\end{equation} 
where $\mathbf{1}(t)$ is 1 if the warmup is end and otherwise 0. 

During inference time, only the prediction of the auxiliary classifier is used to select most likely category.

\section{Experiments}

\subsection{Implement Details}

We first construct imbalance datasets based on several benchmark datasets including CIFAR10, CIFAR100 \cite{krizhevsky2009learning}, STL10 \cite{DBLP:journals/jmlr/CoatesNL11} and SVHN \cite{svhn}. We consider long-tail (LT) distribution as the imbalance type, where the data size for each category decreases exponentially in order. Combined with the imbalance ratio $\gamma$ mentioned in problem setup before, the data number of each class can be expressed as $N_{k}=N_{1}\times\gamma^{-\frac{k-1}{K-1}}$, where $\gamma=\frac{N_{1}}{N_{K}}$ as mentioned before. To construct standard CIFAR10-LT and SVHN-LT dataset for our main experiment, we set $N_{1}=1000$, $\gamma=100$ and $\beta=20\%$. We also try to evaluate our algorithm on a larger imbalance dataset, by following Lee et al. \cite{DBLP:conf/nips/LeeSK21} to set $N_{1}=200$, $\gamma=20$ and $\beta=40\%$ for standard CIFAR100-LT. In building standard STL10-LT dataset, $N_{1}$ is set to 150 and $\gamma=10$ for labeled data, but $\beta$ is not used here since we do not have access to the ground-truth labels of unlabeled data in STL10's training set. 

For base network structure, we follow Lee et al.\cite{DBLP:conf/nips/LeeSK21} to use Wide ResNet-28-2 \cite{DBLP:conf/bmvc/ZagoruykoK16}. The projection head is implemented with a single linear layer of 32-dimension. We implement all the algorithms based on USB \cite{usb2022} framework and use a single RTX 3090 GPU to train models. SGD is used to optimize parameters. Each mini-batch includes 64 labeled samples and 64$\times uratio$ unlabeled samples, and $uratio$ varies for different base SSL algorithms. The learning rate is initially set as $\eta_0=0.03$ with a cosine learning rate decay schedule as $\eta = \eta_0 \cos{(\frac{7 \pi t}{16T})}$. The total number od training for each algorithm is 300,000, and the first 100,000 is warmup stage by default. We report the mean balanced accuracy as well as standard deviation of three trials.

\subsection{Main Results}

We conduct main experiments on standard CIFAR10-LT, CIFAR100-LT, STL10-LT and SVHN-LT datasets. We select FixMatch and ReMixMatch as backbone SSL algorithms and apply several edge-cutting CISSL algorithms including ours over them. The results are reported in Table \ref{table:mainresults}. In each dataset, fully supervised method using only labeled data performs poorly. FixMatch and ReMixMatch take a step forward with the help of using unlabeled data, but there is still a lot of room for improvement. Methods like DASO and DebiasPL make progress on some datasets, but do not perform stably well on others. ABC and CoSSL are typical representatives of existing instance-level method and feature-level method respectively, and they steadily outperform the above mentioned methods. However, it is clear that BaCon achieves new state-of-the-art results across multiple settings. For example, BaCon outperforms CoSSL by 0.37\% on CIFAR10-LT based on FixMatch. Also, on more challenging STL10-LT, FixMatch-based BaCon reaches an accuracy of 71.55\% which exceeds ABC by 0.32\%. On CIFAR100-LT with larger number of categories, our FixMatch version method stably obtains the best result of 57.96\%, which is 0.63\% higher than CoSSL. Based on ReMixMatch, Bacon maintains a gap of 0.53\% on STL10-LT compared with CoSSL and 2.02\% compared with ABC.

\subsection{Different Imbalance Degree}

The imbalance degree of the dataset is a key factor that affects the performance of CISSL algorithms. Methods that have good results under one imbalance degree may fail to maintain them under a steeper distribution. 

Here we evaluate the robustness of our method across different imbalance distribution on CIFAR10-LT and compare with other algorithms. Except for the setting shown in Table \ref{table:mainresults}, a more extreme imbalance ratio $\gamma_{L}=\gamma_{U}=150$ is also implemented. What's more, we also introduce an unusual setting where the imbalance ratios of labeled data and unlabeled data are inversely proportional, to be more specific, $\gamma_{L}=1/\gamma_{U}=100$ and $\gamma_{L}=1/\gamma_{U}=150$. Such an unusual setting may appear counter-intuitive at first glance, but it can test the robustness of the algorithms and whether it strongly relies on prior knowledge of the data distribution. 

We fix $N_1$ to be 4000 as same as in Table \ref{table:mainresults}, and only change $\gamma_{L}$ and $\gamma_{U}$, while other settings remain unchanged. The results are reported in Table \ref{table:imbalancedegree}. A huge drop in performance can be witnessed on CoSSL when it is trained on a inversely proportional dataset. For example, it achieves an accuracy of 83.94\% when $\gamma_{L}=\gamma_{U}=100$, but suffers from a huge decrease of 11.95\% to only 71.99\% when trained on $\gamma_{L}=1/\gamma_{U}=100$. ABC shows a better stability, but leaves with some room for improvement. BaCon not only obtains the best results cross these four settings, but also shows a great robustness against the fluctuation of imbalance degree. For example, BaCon produces an accuracy of 83.80\% on $\gamma_{L}=1/\gamma_{U}=100$ and 82.35\% on $\gamma_{L}=1/\gamma_{U}=150$, which are 2.66\% and 3.51\% higher than the second best.

\begin{table}[!t]
\centering
\tabcolsep=2pt
\renewcommand\arraystretch{1.1}
\resizebox{\columnwidth}{!}{
\begin{tabular}{c c c c c}
\hline \hline 
 \multirow{2}{*}{Method} 
 & $\gamma_{L}=100$ & $\gamma_{L}=100$ & $\gamma_{L}=150$ & $\gamma_{L}=150$ \\ 
  & $\gamma_{U}=100$ & $\gamma_{U}=1/100$ & $\gamma_{U}=150$ & $\gamma_{U}=1/150$\\
\hline
FixMatch & 75.66 & 56.35 & 73.45 & 62.30\\
w/ CReST+ & 79.14 & 66.47 & 74.51 & 62.75\\
w/ ABC & 82.48 & 81.14 & 79.41 & 78.84\\
w/ CoSSL & 83.94 & 71.99 & 81.83 & 74.14\\
\rowcolor{lightgray} \textbf{w/ BaCon} & \textbf{84.61} & \textbf{83.80} & \textbf{81.99}  & \textbf{82.35}\\
\hline \hline
\end{tabular}}
\caption{Ablation studies on different imbalance degree. $\gamma_{L}$ and $\gamma_{U}$ represents the imbalance ratio of labeled and unlabeled data respectively. CReST+: CReST+PDA.}
\label{table:imbalancedegree}
\end{table}

\begin{table}[!t]
\centering
\resizebox{\columnwidth}{!}{
\begin{tabular}{c c c c|c}
\hline \hline
Contra &RNS &Naive BTA &Decay BTA &Acc(\%)\\
\hline
$\checkmark$  &$\checkmark$ &  &  &84.30 \\
$\checkmark$  & &$\checkmark$  &  &83.87 \\
$\checkmark$  &$\checkmark$ &$\checkmark$  &  &84.15 \\
$\checkmark$  &$\checkmark$ &  &$\checkmark$  &\textbf{84.61} \\
\hline \hline
\end{tabular}}
\caption{Ablation for proposed components. Contra: Contrastive loss. Naive BTA: BTA  without iteration decay.}
\label{table:mainablation}
\end{table}

\begin{table}[!t]
\centering
\renewcommand\arraystretch{1.1}
\resizebox{\columnwidth}{!}{
\begin{tabular}{c|c c c c c c}
\hline \hline
 &Identity &Nonlinear &32-D &128-D &512-D \\
\hline 
Acc(\%) &82.86 &83.95 &\textbf{84.61}  &83.48  &82.64 \\
\hline \hline
\end{tabular}}
\caption{Ablation studies of projection mode for contrastive learning space. Identity: identity mapping of representation. }
\label{table:projablation}
\end{table}

\begin{figure}[t!]
    \begin{subfigure}[t]{0.155\textwidth}
           \centering
           \includegraphics[width=\textwidth]{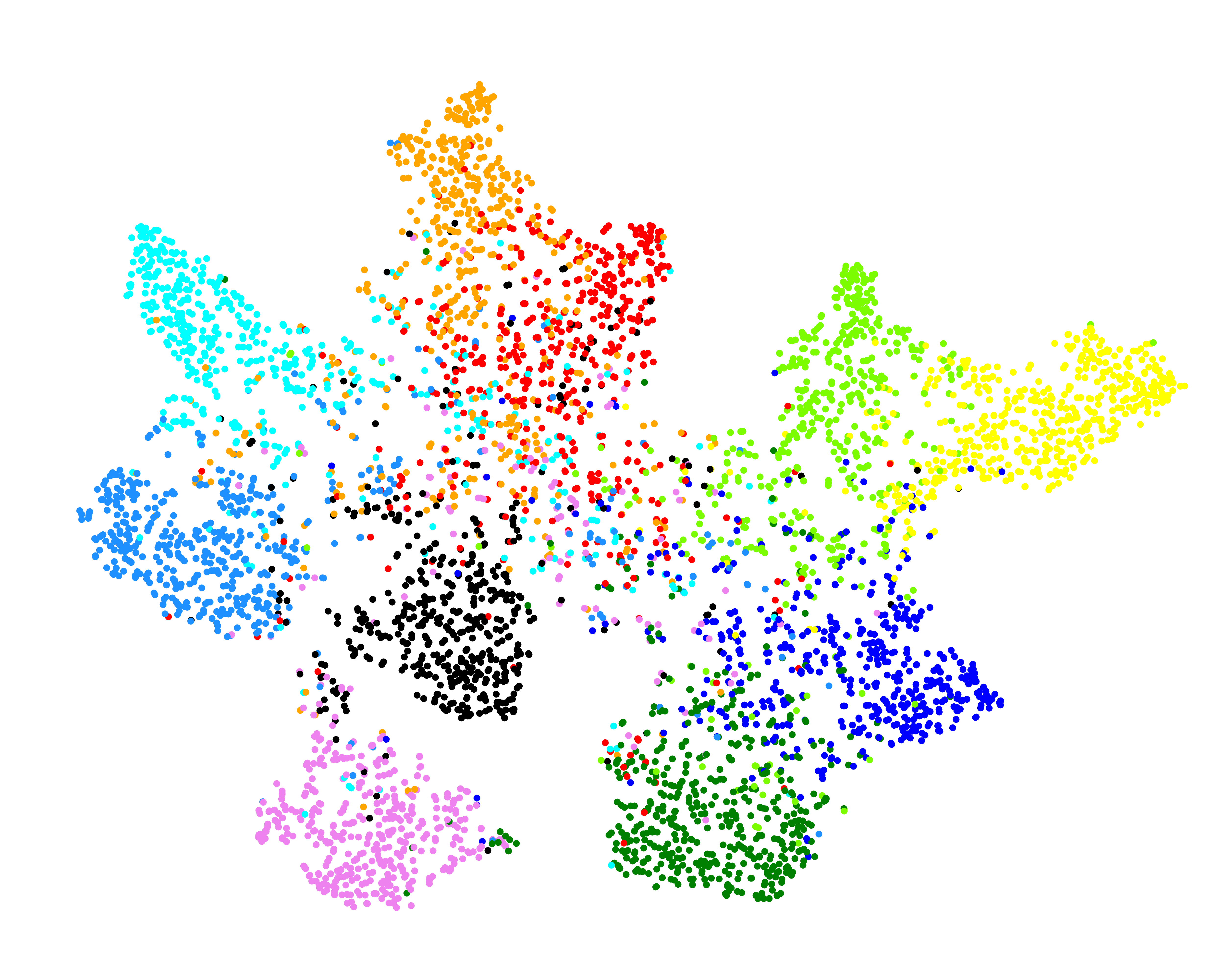}
            \caption{FixMatch}
            \label{fig:tsne-fix}
    \end{subfigure}
    \begin{subfigure}[t]{0.155\textwidth}
            \centering
            \includegraphics[width=\textwidth]{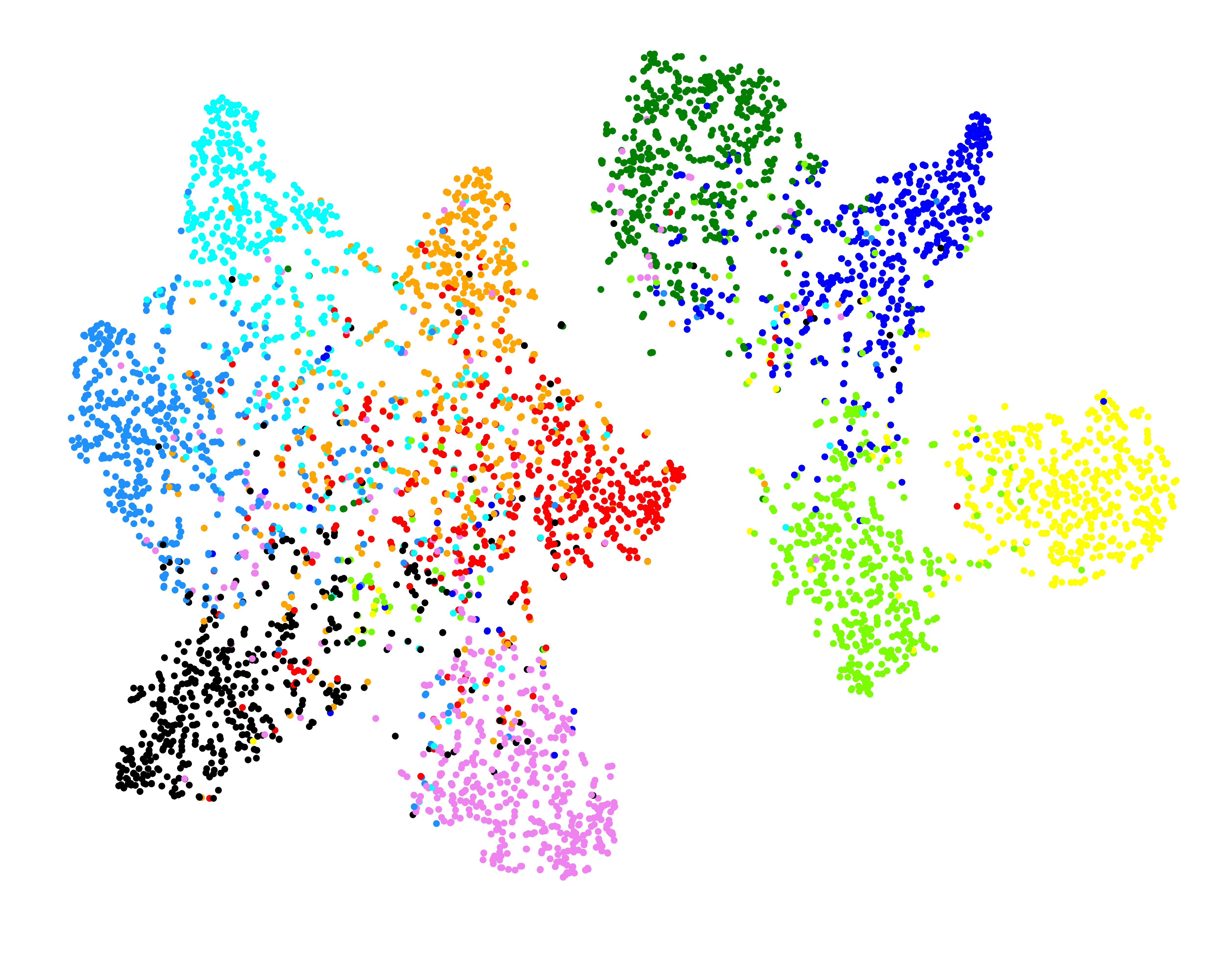}
            \caption{FixMatch+ABC}
            \label{fig:tsne-abc}
    \end{subfigure}
    \begin{subfigure}[t]{0.155\textwidth}
            \centering
            \includegraphics[width=\textwidth]{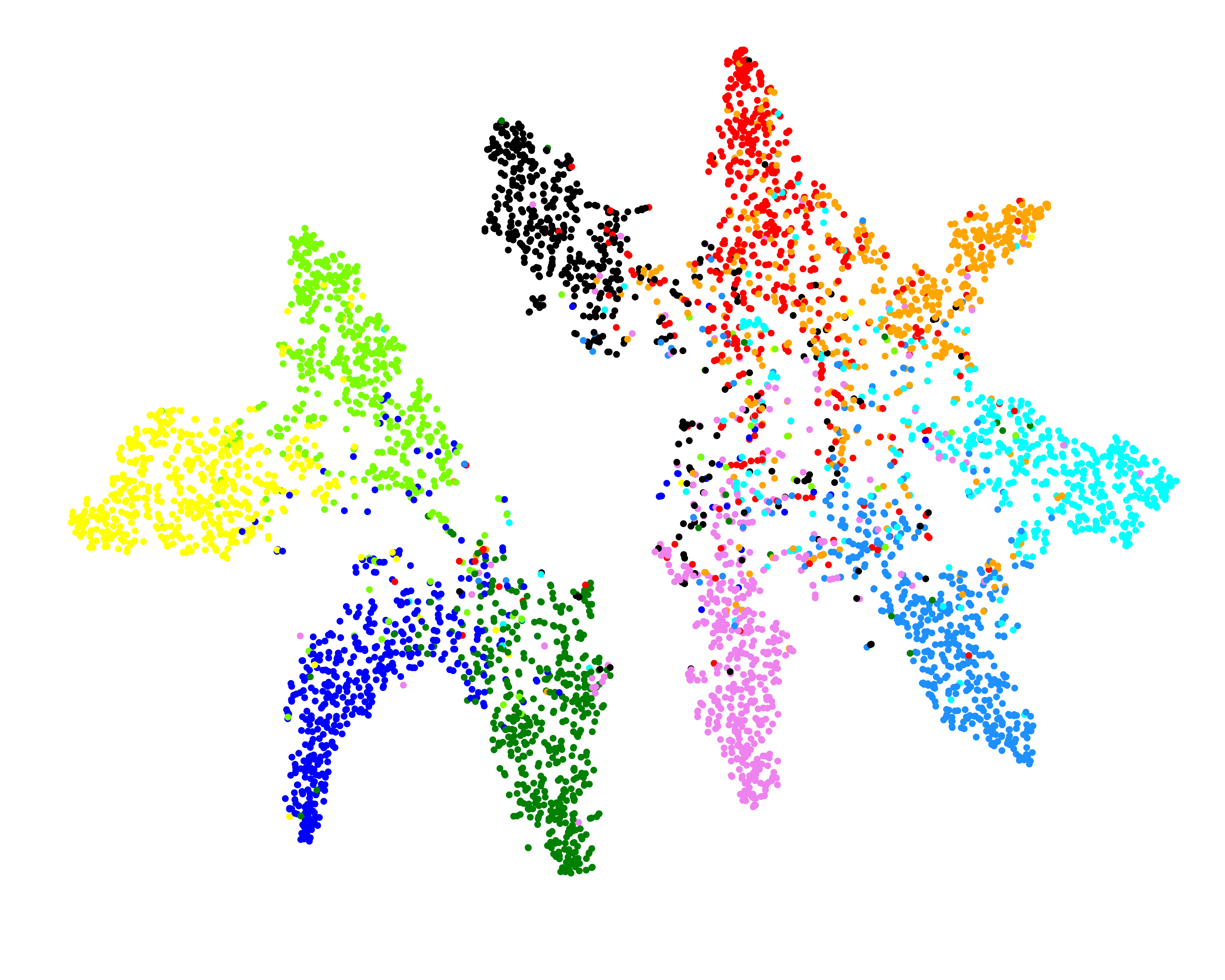}
            \caption{Ours}
            \label{fig:tsne-bacon}
    \end{subfigure}
    \caption{t-SNE visualization results of balanced test set learned by algorithms trained on CIFAR10-LT.}
    \label{fig:tsne}
\end{figure}

\subsection{Ablation Studies}

To evaluate the effectiveness of each component we propose in BaCon, we conduct a series of ablation studies on standard CIFAR10-LT based on FixMatch. 

We first test the performance of backbone SSL with only an auxiliary classifier attached to it, and then add each component one by one. The results are shown in Table \ref{table:mainablation}. Baseline with the auxiliary classifier achieves an accuracy of 83.95\%, and with a contrastive loss that uses RNS the accuracy raises to 84.30\%. However, a naive BTA with no iteration decay brings no gain, and the accuracy drops to 83.87\% after RNS is abandoned. Finally, when the iteration decay is used in BTA, we witness the best performance of 84.61\%. These experiments can verify the effectiveness of our proposed methods.

In addition, we further explore different projection method towards target contrastive space. As comparisons, we choose identity projection, linear projections of different mapping dimensions, and a nonlinear layer constructed by adding $ReLU$ function to the 32-D linear projection. The results are shown in Table \ref{table:projablation}. It is obvious that the identity projection largely brings down the performance and 32-D linear projection obtains the highest result. Nonlinear projection does not perform well here, and we argue that this is because nonlinear layers can filter out task-related information, and it is more likely to be useful in pretraining task as in \cite{chen2020simple}.

\subsection{Qualitative Analysis}

BaCon builds more balanced representations for downstream classification. To verify this, we present the visualization of t-distributed stochastic neighbor embedding (t-SNE) \cite{van2008visualizing} for the representations of the balanced CIFAR-10 test set learned by FixMatch, FixMatch-based ABC and FixMatch-based Bacon respectively, and the results are shown in Figure \ref{fig:tsne}. Normal SSL algorithm FixMatch fails to produce separable boundaries. With distribution-related 0/1 mask, ABC presents clearer boundaries than FixMatch, but there is still a great deal of confusion among most categories. BaCon directly implements feature-level clustering, and produces more separable representations. This further proves the gains of our feature-level approach.

\section{Conclusion}

In this paper, we discuss the limitation of instance-level CISSL method with biased representation. Next, we propose our feature-level contrastive learning method BaCon to deal with this problem. BaCon projects backbone's representation to another feature space and computes class-wise feature centers as positive anchors. In addition, the proposed RNS method is used to efficiently find sufficient reliable negative samples. Considering the imbalanced distribution, a balanced temperature regulation mechanism is also designed. Finally we show that extensive experiments have verified the effectiveness of our method.

\section{Acknowledgments}

This work was supported by National Key R\&D Program of China (No.2018AAA0100300) and the funding of China Tower.

\bibliography{aaai24}

\end{document}